\documentclass[10pt,twocolumn,letterpaper]{article}

\usepackage{cvpr}              

\usepackage[dvipsnames]{xcolor}

\usepackage{bbm}
\usepackage{multirow}
\usepackage{color, colortbl}
\usepackage[accsupp]{axessibility}

\definecolor{lightgray}{gray}{0.92}
\definecolor{cvprblue}{rgb}{0.21,0.49,0.74}
\usepackage[pagebackref,breaklinks,colorlinks,citecolor=cvprblue]{hyperref}

\title{HumMUSS: Human Motion Understanding using State Space Models}

\author{Arnab Mondal\\
Mila \& Apple\\
{\tt\small arnab.mondal@mila.quebec}
\and
Stefano Alletto\\
Apple\\
{\tt\small salletto@apple.com}
\and
Denis Tome\\
Apple\\
{\tt\small d\_tome@apple.com}
}

\begin{document}
\twocolumn[{
\renewcommand\twocolumn[1][]{#1}
\maketitle
\begin{center}
\includegraphics[width=0.9\linewidth]{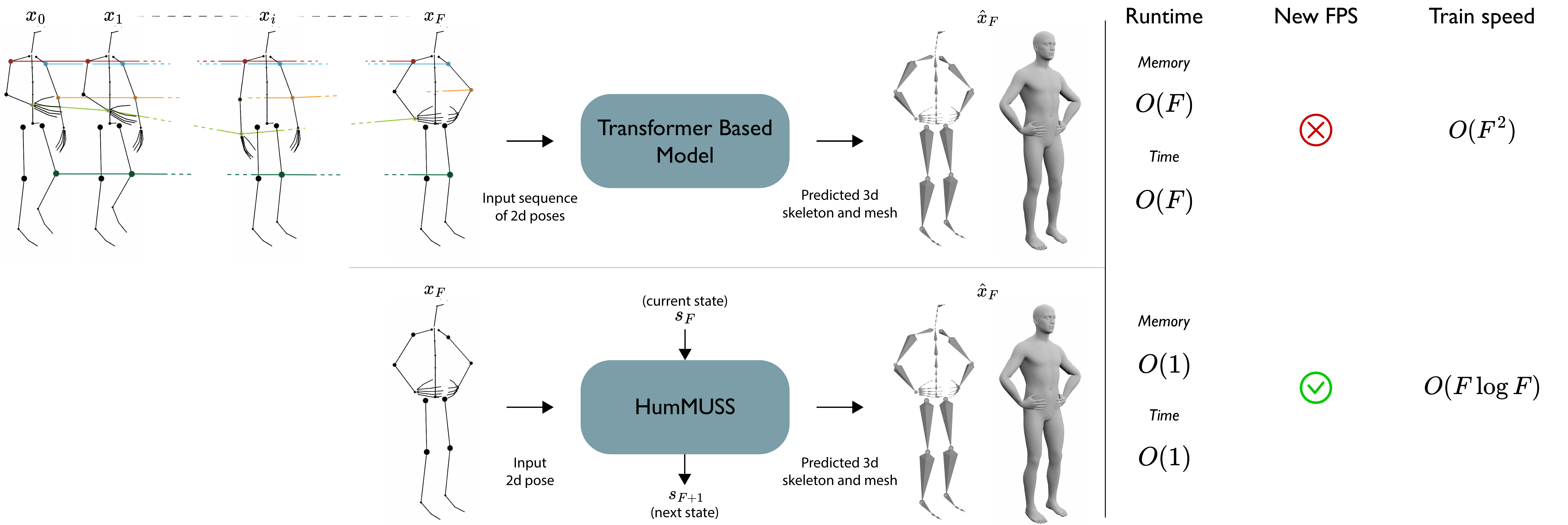}
\captionof{figure}{HumMUSS vs. Transformer-Based Models for sequential prediction of 3D poses and human meshes from 2D keypoint videos. \emph{\bf Top}: Transformer-based models attend to a history of 2D poses/keypoints to predict the current frame's output. \emph{\bf Bottom}: HumMUSS, being a stateful model, efficiently utilizes only the current frame and its current state for predictions, ensuring constant memory and time complexity. HumMUSS also generalizes to new frame rates and enhances the training speed without compromising the prediction accuracy. 
}
\vspace{1mm}
\end{center}
}]

\begin{abstract}
Understanding human motion from video is essential for a range of applications, including pose estimation, mesh recovery and action recognition. While state-of-the-art methods predominantly rely on transformer-based architectures, these approaches have limitations in practical scenarios. Transformers are slower when sequentially predicting on a continuous stream of frames in real-time, and do not generalize to new frame rates. In light of these constraints, we propose a novel attention-free spatiotemporal model for human motion understanding building upon recent advancements in state space models.

Our model not only matches the performance of transformer-based models in various motion understanding tasks but also brings added benefits like adaptability to different video frame rates and enhanced training speed when working with longer sequences of keypoints. Moreover, the proposed model supports both offline and real-time applications. For real-time sequential prediction, our model is both memory efficient and several times faster than transformer-based approaches while maintaining their high accuracy.

\end{abstract}
\section{Introduction}
\label{sec:intro}
Understanding human motion is crucial for various computer vision applications, including body keypoints tracking~\cite{chun2023learnable, iskakov2019learnable, tang20233d, zhu2023motionbert}, human mesh estimation~\cite{choi2020pose2mesh,zhang2021pymaf} or action recognition~\cite{zhou2023learning,duan2023skeletr}. While recent advancements  have enabled a plethora of real-world applications focusing on real-time requirements and mobile deployment~\cite{bazarevsky2020blazepose,yao2023poserac}, this field is dominated by Transformer-based architectures~\cite{yao2023poserac,li2023token,zhu2023motionbert}. Although these models attain remarkable accuracy, they face practical challenges: processing long video sequences with Transformers can be slow, and they do not generalize well to unseen frame rates. Moreover, Transformers are highly inefficient in terms of memory and speed for real-time sequential prediction~\cite{sun2023retentive,liu2023efficientvit,pope2023efficiently} compared to state-based sequence models. 


In this work, we propose to look beyond attention-based architectures for learning motion representations and leverage recent advancements in State Space Models (SSMs) \cite{gu2022efficiently, gu2022parameterization, gupta2022diagonal, mehta2022long, wang2022pretraining} to address the practical limitations of Transformers.
In particular, we propose a novel attention-free spatiotemporal architecture, named HumMUSS, which utilizes SSMs to learn human motion representations. 
HumMUSS consists of several stacked blocks, each containing two streams of alternating spatial and temporal Gated Diagonal SSM blocks (GDSSM), designed to efficiently learn rich spatiotemporal features.
HumMUSS inherits the advantages of DSSM, such as faster training and inference for longer sequences and O(1) time and memory complexity for real-time sequential inference. Moreover, it also can generalize to unseen and variable frame rates due to SSMs' inherent continuous time formulation.
This is especially relevant for any real-time on-device applications where load and thermal conditions can drastically change the camera's sampling rate.

HumMUSS achieves performance on par with state-of-the-art methods 
on several motion understanding tasks, while providing all the aforementioned practical benefits. Additionally, we introduce a fully causal version of HumMUSS, designed to predict each current time-step using only past and current frames, with no foresight into future frames.
In this causal setting, we demonstrate that HumMUSS not only outperforms current state-of-the-art models but is also several times faster and memory efficient. This is essential for real-time applications where low latency is critical.

Our contributions can be summarized as follows
\begin{itemize}
    \item 
    We introduce HumMUSS, a novel attention-free spatiotemporal architecture using SSMs, to process human motion.
    To the best of our knowledge, this is the first attempt to bring SSM-based architectures to human motion understanding tasks.
    \item We demonstrate HumMUSS improves both training and inference speed compared to state of the art methods such as MotionBERT~\cite{zhu2023motionbert} for long motion videos. Additionally, being a state-based model, it makes sequential prediction several times faster, for example $3.8\times$ for a context length of $\approx 243$ frames.
    \item We show that HumMUSS, being a continuous-time model, seamlessly generalizes to dynamic frame rates during inference with minimal performance degradation.
    \item We empirically demonstrate that our pre-trained HumMUSS achieves competitive accuracy for various tasks such as 3D pose estimation, human mesh recovery and action recognition, proving its viability as a superior alternative to transformers for a broad range of motion understanding tasks.
\end{itemize}
\section{Related Work}
\label{sec:related_work}


\paragraph{3D Pose Estimation}
Various approaches for Human 3D Pose Estimation are currently implemented using either a monocular (single-view) or a multi-view configuration. In the multi-view scenarios~\cite{tome2018rethinking, iskakov2019learnable, remelli2020lightweight, reddy2021tessetrack, zhang2021adafuse, chun2023learnable}, multiple camera perspectives are utilized to enhance the accuracy of the extracted poses by exploiting geometric information from the camera placement. Both configurations can be further classified into two broader categories: \emph{i)} direct approaches, which predict 3d joints directly from input frames, and \emph{ii)} two-step solutions, which use 2d joint positions as a basis to estimate the 3d poses.

Direct 3D estimation techniques~\cite{pavlakos2018ordinal, pavlakos2017coarse, sun2018integral, zhou2019hemlets} determine the spatial positions of 3D joints from video frames without intermediary stages, enabling real-time application with reduced computational overhead. In comparison to that, other 2D-3D lifting techniques have adopted an initial step of utilizing a readily available precise off-the-shelf 2D pose estimators~\cite{wang2018cascaded, newell2016stacked, sun2019deep}, and then lift the 2D coordinates into 3D~\cite{tang20233d, zhang2022mixste, zhao2023poseformerv2, zheng20213d, tome2017lifting, zhu2023motionbert, li2022mhformer, pavllo20193d, zheng20213d, martinez2017simple}. 

\paragraph{Human Mesh Recovery}
Given a frame or video input, the problem of 3D human mesh recovery falls into two main categories: parametric and non-parametric methods. Parametric methods learn to estimate the parameters of a human body model, like SMPL~\cite{loper2023smpl}, while non-parametric approaches directly regress the 3D coordinates of human mesh vertices.  
Parametric models~\cite{bogo2016keep, dwivedi2021learning, guan2009estimating, jiang2020coherent, hmrKanazawa17, kocabas2021pare, kolotouros2019spin, pavlakos2018learning, sun2021monocular, zhang2021pymaf} utilizing the inherent body structure encoded in models such as SMPL~\cite{loper2023smpl}, estimate shape and pose parameters, which often lead to more anatomically accurate results. 
However, as the 3d position of vertices are generated from shape and pose parameters, such methods can sometimes limit the representation of complex and diverse body types and movements.
In contrast, non-parametric models~\cite{choi2020pose2mesh, kolotouros2019convolutional, lin2021end, lin2021mesh, moon2020i2l} capture more nuanced details and variations in human forms and postures, offering adaptability to a wider range of complex data. However, they may yield results that lack interpretability due to the absence of explicit parameters, which can obscure the rationale behind certain predictions.


\paragraph{Skeleton-based Action Recognition}
Skeleton-based action recognition focuses on identifying the spatial and temporal dynamics between human body joints to classify the action being performed in a video.
 Recent methods approach this challenge by employing multi-task learning or using multimodal inputs \cite{ahn2023star,luvizon20182d}, yet the predominant strategy involves learning from pre-established 2D or 3D poses \cite{zhou2023learning,duan2023skeletr,foo2023unified,song2017end,chen2021channel,duan2022revisiting}. 
Among these methods, some of the most relevant approaches rely on one three main paradigms: 3D CNNs, GCNs or transformers. \cite{duan2022revisiting} utilizes 3D CNNs alongside 3D volumetric heatmaps to learn a representation that is robust to noise in the input pose.
\cite{zhou2023learning} proposes an additional feature refinement phase to mitigate some intrinsic limitations of graph-based pose modeling with GCNs.
Lastly, 
\cite{zhu2023motionbert} leverages a Transformer, pre-trained on a pretext 3D pose estimation task, to classify actions from 2D pose data.

\paragraph{Transformers for Human Modeling}
The Transformer architecture~\cite{vaswani2017attention}, initially developed for language modeling, has demonstrated significant advancements in various computer vision tasks. Particularly in human understanding tasks, the Transformer has shown notable improvements in the accuracy of human pose and shape estimation~\cite{kocabas2021pare, lin2021end, shi2022end, wan2021encoder, yuan2022glamr}. Several approaches utilize the attention mechanism to effectively capture spatial relationship of the joints~\cite{kocabas2021pare, lin2021mesh, lin2021end} and their temporal dynamics~\cite{wan2021encoder, yuan2022glamr}, or a combination of both using spatiotemporal attention based architectures~\cite{li2022mhformer, li2021tokenpose, zhang2022mixste, zhu2023motionbert, zhao2023poseformerv2}. These methods indicate the versatility and effectiveness of Transformers in modeling complex human bodies and their motion data.



\paragraph{State Space Models}

Recent studies~\cite{gu2022efficiently, gu2022parameterization, gupta2022diagonal, mehta2022long} have explored State Space Models (SSMs) as an effective alternative to Transformers for modeling long sequences. In particular, the use of a diagonal state matrix has been crucial in enhancing the training speed of these models, thus making them faster than Transformers, especially for processing longer sequences. Moreover, certain initializations of diagonal values~\cite{gu2022parameterization} are capable of replicating convolution kernels that are designed for long-range memory. This aligns with the principles of the HiPPO theory~\cite{gu2020hippo} and explains the comparable performance of Diagonal SSMs to Structured State Space Models (S4)~\cite{gu2022efficiently} as indicated by~\cite{gupta2022diagonal}. 

\paragraph{Multiplicative Gating}
Various types of neural architectures, including Multi-Layer Perceptrons (MLPs), CNNs, and Transformers have benefited from the integration of gating units~\cite{dauphin2017language, shazeer2020glu}. One popular form of these gating units is the Gated Linear Unit (GLU), which has proven particularly effective in CNN-based Natural Language Processing (NLP) applications~\cite{dauphin2017language}. 
Recent studies emphasize the utility of gating units for simplifying network routing and suggests that multiplicative gating can act as a surrogate to recapture some of the intricate interactions found in attention-based mechanisms. For instance, Hua~\etal~\cite{hua2022transformer} demonstrate that linear-time attention models can achieve better performance with the use of optimized gating mechanisms. Similarly, \cite{mehta2022long} and~\cite{wang2022pretraining} show that incorporating multiplicative gating mechanism enhances SSMs for NLP tasks.
\section{Background}
\label{sec:background}

\subsection{State Space Models}
State space models (SSM) provide a continuous time learnable framework for mapping between continuous-time scalar inputs and outputs. Given an input \( u(t) \) and output \( y(t) \), SSM is described by differential equations involving a continuous-time state vector \( x(t) \) and its derivative \( x'(t) \), parameterized by matrices \( A \in \mathbb{R}^{N \times N} \), \( B \in \mathbb{R}^{N \times 1} \), \( C \in \mathbb{R}^{1 \times N} \) and \( D \in \mathbb{R} \).
\begin{align*}
x'(t) = A x(t) + B u(t),  \quad y(t) = C x(t) + D u(t).
\end{align*}

In discrete-time, with a parameterized sample time $\Delta$, these equations transition into recursive formulations
\begin{align}
x_k = \bar{A} x_{k-1} + \bar{B} u_k, \quad y_k = \bar{C} x_k + \bar{D} u_k.
\label{eqn:ssm_recursive}
\end{align}

where $ \bar{A} = e^{A \Delta}$, $\bar{B} = (e^{A \Delta} - I)A^{-1}B$, $\bar{C} = C$ and $\bar{D} = D$ using zero order hold~(zoh) discretization~\cite{iserles2009first}.

The linear nature of SSMs allows the output sequence to be computed directly by unrolling the recursion in time
\begin{align*}
y_k = \sum_{j=0}^{k} \bar{C} \bar{A}^j \bar{B} \cdot u_{k-j}.
\end{align*}
A significant advantage of this structure is the potential for parallel computation, facilitated by the discrete convolution of the input sequence \( u \) with the precomputed SSM kernel \( K = (\bar{C}\bar{B}, \bar{C}\bar{A}\bar{B}, \dots, \bar{C}\bar{A}^{L-1}\bar{B})\), denoted by $y = K \ast_c u$.
While the naive approach to this computation requires \( O(L^2) \) multiplications, it can be done in \( O(L \log(L)) \) time using the Fast Fourier Transform (FFT) \cite{brigham1988fast}. SSMs can conveniently switch from their convolutional to recursive formulation in~\cref{eqn:ssm_recursive} when properties like auto-regressive decoding are desirable.  

\subsection{Diagonal State Spaces}
An efficient adaptation of the SSM framework is the incorporation of a diagonal state matrix, greatly facilitating the computation of the SSM kernel \( K \)~\cite{gu2021efficiently, gupta2022diagonal}. As shown by Gu~\etal~\cite{gu2022parameterization}, a diagonal state matrix \( A \) represented as \mbox{$\Lambda = \text{diag}(\lambda_1, \ldots, \lambda_N)$} can approximate the HiPPO parameterization~\cite{gu2020hippo} of the transition matrix A that yields stable training regime with long sequences. Further simplifications are introduced with the vector \( B \) being expressed as \mbox{$B=(1)_{N\times 1}$}. Under these conditions, the DSSM model is characterized by learnable parameters \mbox{$\Lambda_{re},\Lambda_{im} \in \mathbb{R}^N$}, \mbox{$C \in \mathbb{C}^N$}, and \mbox{$\Delta_{\log} \in \mathbb{R}$}. The diagonal elements of \( A \) are then computed through the relationship \mbox{$-\exp(\Lambda_{re}) + i \cdot \Lambda_{im}$}, where $i = \sqrt{-1}$ and $\Delta$ is deduced as \mbox{$\text{exp}(\Delta_{log}) \in \mathbb{R}^{>0}$}. The kernel $K$ can be computed as
\begin{equation}
    K = (C \odot \begin{bmatrix}\left(e^{\lambda_1 \Delta} - 1\right)/\lambda_1\\ \vdots \\\left(e^{\lambda_N \Delta} - 1\right)/\lambda_N\end{bmatrix})^T \cdot \text{exp}(P)
\label{eqn:eff_kernel}
\end{equation} 
where $\odot$ is element-wise multiplication and the elements of matrix \mbox{$P \in \mathbb{C}^{N\times L}$} are defined as \mbox{$P_{j,k} = \lambda_j k \Delta$}. In practice, to get a real valued kernel K, the diagonal elements are assumed to appear in complex conjugate pairs and their corresponding parameters in $C$ are tied together. Hence, the dimension of state space is effectively set to $N/2$ and the final kernel is obtained by taking the real part of $2K$. We provide more details on the initialization of DSSM parameters in the supplementary material.

This framework establishes a linear mapping for 1-D sequences. When extending to sequences comprising $H$-dimensional vectors, individual state space models are applied to each of the $H$ dimensions. Specifically, a DSSM layer takes a sequence of length $L$, denoted as $\mathbf{u} \in \mathbb{R}^{H \times L}$, and yields an output $\mathbf{y} \in \mathbb{R}^{H \times L}$. For each feature dimension $h = [1, \dots, H]$, a kernel $K_h \in \mathbb{R}^L$ is computed. The corresponding output $y_h \in \mathbb{R}^L$ for this feature is obtained using the convolution of input $u_h \in \mathbb{R}^L$ and kernel $K_h$. This can be done for a batch of samples leading to a linear DSSM layer that can map from  $\mathbf{u} \in \mathbb{R}^{B \times L \times H}$ to $\mathbf{y} \in \mathbb{R}^{B \times L \times H}$ and is denoted by \mbox{$y = \text{DSSM}(u)$}\footnote{The transpose operation of the last two dimension of the input and output is moved inside the DSSM layer.}.
Considering a batch size of $B$, sequence length $L$, and hidden dimension $H$, the computation time for the kernels in the DSS layer scales as $O(NHL)$, whereas the discrete convolution demands a time complexity of $O(BHL \log(L))$.

\section{Method}
\label{sec:method}

\begin{figure}[t]
    \centering
    \includegraphics[width=.95\columnwidth]{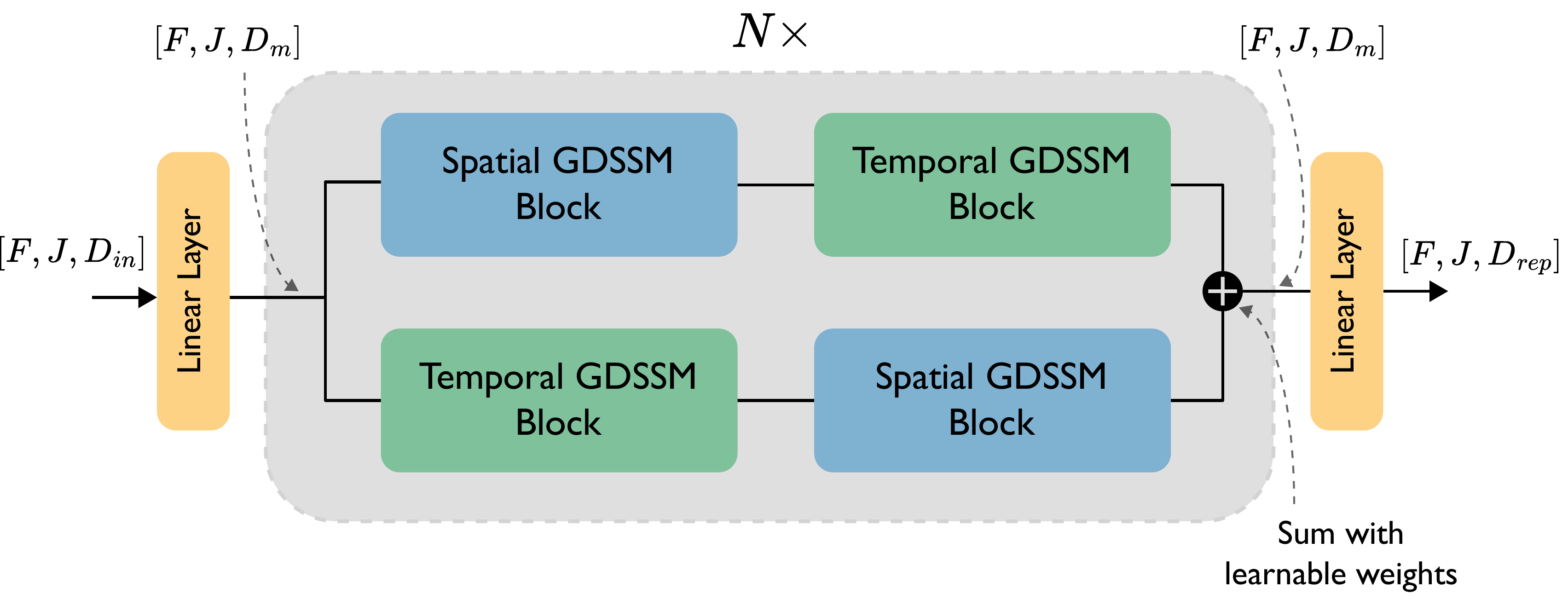}
    \caption{HumMUSS model architecture}
    \label{fig:hummus}
\end{figure}

HumMUSS is a general purpose attention-free architecture that takes spatiotemporal human motion data as input and outputs representations corresponding to each spatial and temporal location. Let the input to HumMUSS be a video of joints $u \in \mathbb{R}^{B \times F \times J\times D_{in}}$ where $B$ is the batch size, $F$ is the number of frames, $J$ is the number of joints and $D_{in}$ is the dimension of the input which is typically $3$ for the 2D joint positions and scalar joint confidence. HumMUSS learns to model the underlying continuous signal resulting from evolution of joint positions and their interactions with each other to produce a spatiotemporal representation $r \in \mathbb{R}^{B \times F \times J\times D_{rep}}$.  As shown in Figure \ref{fig:hummus}, HumMUSS features a sequence of spatiotemporal blocks, each incorporating two alternating streams of Spatial and Temporal Gated Diagonal State Space Model (GDSSM) blocks. It uses a lifting layer to transform the input to the model dimension $D_{m}$ and a final layer to transform the output embeddings in $D_{m}$ to required representation dimension $D_{rep}$.

In the subsequent sections, we first provide our general architecture of a bidirectional and uni-direction (causal) GDSSM blocks, and then show how to use them to build a spatiotemporal layer.

\subsection{Bidirectional GDSSM Block}
\label{subsec:bigdssm}
\begin{figure}[t]
    \centering
    \includegraphics[width=0.95\columnwidth]{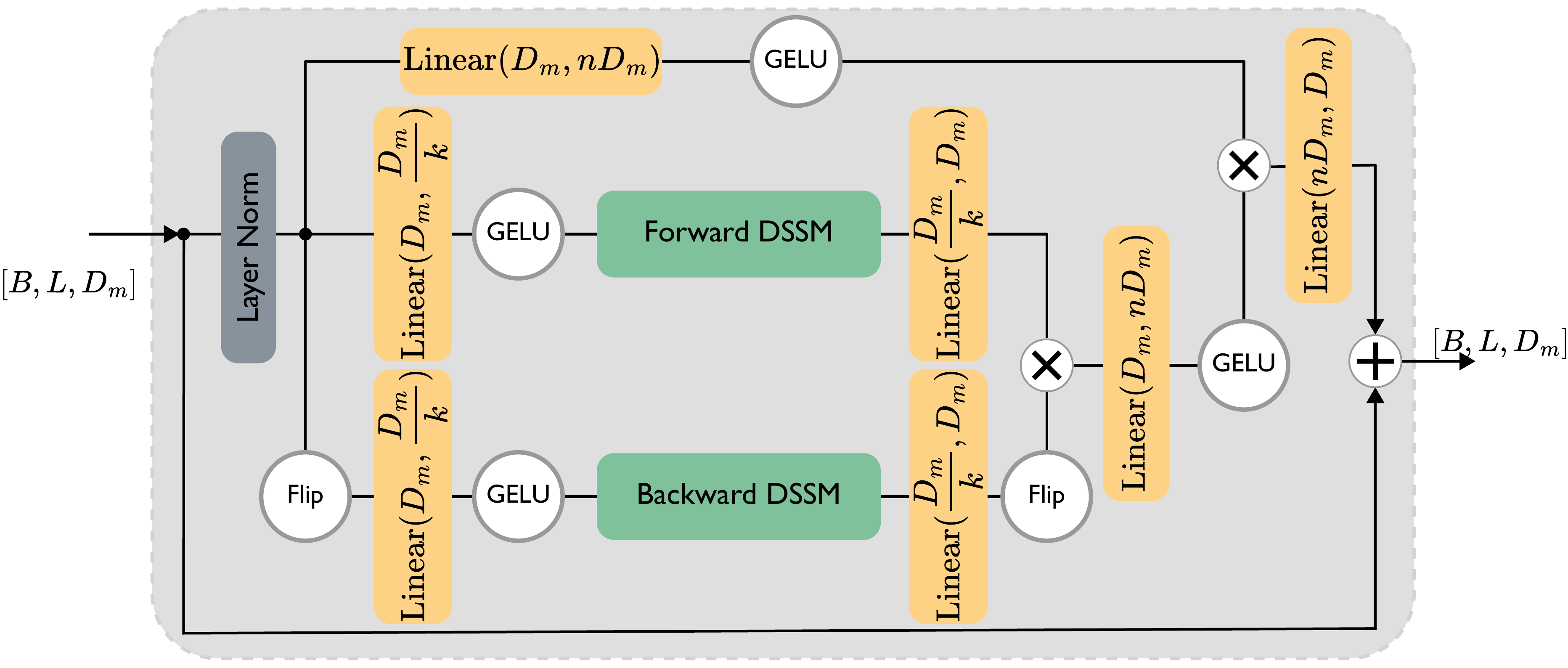}
    \caption{Bidirectional Gated DSSM Block}
    \label{fig:hummus_block}
    \vspace{-3mm}
\end{figure}

Given an input $x \in \mathbb{R}^{B \times L\times D_{m}}$ where $B$ is the batch size, $L$ is the length of the sequence, and $D_{m}$ is the model dimension, the Bidirectional GDSSM Blocks learns to aggregate information across the sequence dimension $L$. As illustrated in Figure~\ref{fig:hummus_block}, this block starts with a layer normalization and has three distinct pathways to process information. The first pathway processes information independently whereas the other two combines it forward and backward in the sequence dimension
\begin{align*}
&x_{id}=\sigma(\text{LayerNorm}(x) W_{id}) &\in &\mathbb{R}^{B \times L\times nD_{m}} \\
&x_f=\text{DSSM}_{f}(\sigma(x_N W_{f}^1))W_{f}^2 &\in &\mathbb{R}^{B \times L\times D_{m}} \\
&x_b=flip(\text{DSSM}_{b}(\sigma(flip(x_N) W_{b}^1))W_{b}^2) &\in &\mathbb{R}^{B \times L\times D_{m}}
\end{align*}

with $W_{id} \in \mathbb{R}^{D_m\times nD_{m}}$, $W_{f}^1-W_{b}^1 \in \mathbb{R}^{D_m\times \frac{D_{m}}{k}}$, $W_{f}^2 - W_{b}^2 \in \mathbb{R}^{ \frac{D_{m}}{k}\times D_m}$ the learnable weight matrices; $flip(\cdot)$ denotes flipping operation along the sequence dimension and $\sigma(\cdot)$ denote GELU activation \cite{hendrycks2016gaussian}. In this formulation, we reduce the dimension of the DSSM by a factor of $k$ to speed up kernel computation and combine different dimensions of the DSSM output by using weights $W_{f}^2, W_{b}^2$. Following~\cite{mehta2022long} and~\cite{wang2022pretraining}, we combine the forward and backward aggregated information using multiplicative gating
\begin{align*}
x_{cb} = \sigma((x_f\odot x_b) W_{cb}) \quad \in \mathbb{R}^{B \times L\times nD_{m}}
\end{align*}
where $W_{cb} \in \mathbb{R}^{D_m\times nD_{m}}$ and $\odot$ denotes a Hadamard Product. Finally, the block's output is computed by combining the independently processed information from the first pathway with the outputs of the other two pathways, and then adding a skip connection with the block's input.
We use a dimension expansion factor of $n$ before using the multiplicative gating
\begin{align*}
x_{out} = x + (x_{cb} \odot x_{id})W_{out} \quad \in \mathbb{R}^{B \times L\times D_{m}}
\end{align*}
where $W_{out} \in \mathbb{R}^{nD_m\times D_{m}}$ is used to bring the output of this multiplicative gate back to the model dimension. This provides an expressive non-linear bidirectional block to process a sequence of vectors and we denote it as \mbox{$x_{out}=\text{BiGDSSM-Block}(x)$}.

\subsection{Unidirectional GDSSM Block}
\label{subsec:unigdssm}
\begin{figure}[t]
    \centering
    \includegraphics[width=0.95\columnwidth]{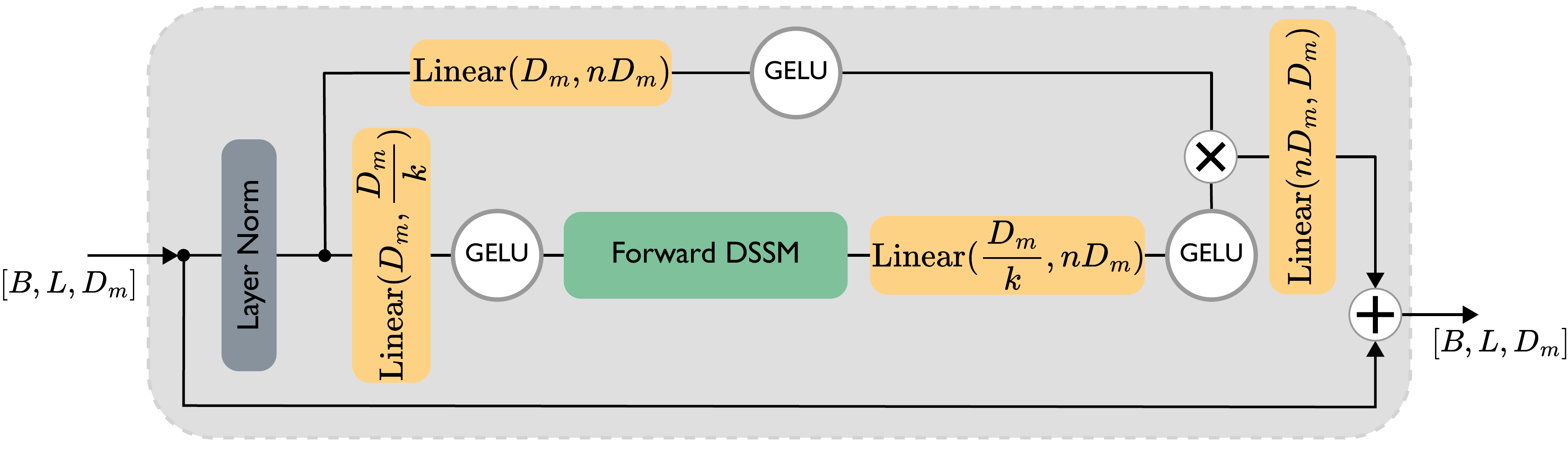}
    \caption{Unidirectional Gated DSSM Block}
    \label{fig:hummus_block_uni}
    \vspace{-3mm}
\end{figure}

Similarly, for an input $x \in \mathbb{R}^{B \times L\times D_{m}}$, the unidirectional GDSSM Blocks learns to combine information along the sequence dimension $L$ but only in forward direction. As shown in Figure \ref{fig:hummus_block_uni}, this block also starts with layer normalization and has two main pathways to process information. The first pathway processes information independently whereas the other one combines it forward in the sequence dimension
\begin{align*}
&x_{id}=\sigma(\text{LayerNorm}(x) W_{id}) &\in &\mathbb{R}^{B \times L\times nD_{m}}\\
&x_f=\text{DSSM}_{f}(\sigma(x_N W_{f}^1))W_{f}^2 &\in &\mathbb{R}^{B \times L\times D_{m}}
\end{align*}
where $W_{id} \in \mathbb{R}^{D_m\times nD_{m}}$, $W_{f}^1 \in \mathbb{R}^{D_m\times \frac{D_{m}}{k}}$ and $W_{f}^2 \in \mathbb{R}^{ \frac{D_{m}}{k}\times nD_m}$. In contrast to the bidirectional block, the output of the unidirectional block is directly computed by combining $x_{id}$ and $x_f$ using multiplicative gating and a skip connection with the input to the block
\begin{align*}
x_{out} = x + (x_{f} \odot x_{id})W_{out} \quad \in \mathbb{R}^{B \times L\times D_{m}}
\end{align*}
where  $W_{out} \in \mathbb{R}^{nD_m\times D_{m}}$. We denote this causal block as $x_{out} = \text{UniGDSSM-Block}(x)$.

\subsection{Building a spatiotemporal Layer}
\label{subsec:spatiotemporal}
In this section, we construct a spatiotemporal layer utilizing the GDSSM Blocks discussed previously. Following~\cite{zhu2023motionbert}, given input $x \in \mathbb{R}^{B \times F \times J \times D_{m}}$, we pass it through two different information processing streams. By adopting this approach, each stream captures distinct spatio-temporal aspects, thereby enhancing the overall expressivity of the model. The first stream combines information spatially and then temporally
\begin{align*}
 x_s &= \text{BiGDSSM-Block}_s^1(x.\text{flatten}(0,1))\\
 x_s &= x_s.\text{reshape}(B, F, J, D_m).\text{T}(1,2)\\
 x_{ts} &= \text{BiGDSSM-Block}_t^1(x_s.\text{flatten}(0,1))\\
 x_{ts} &= x_{ts}.\text{reshape}(B, J, F, D_m).\text{T}(1,2) 
\end{align*}
where $\text{BiGDSSM\-Block}^1_s(\cdot)$-$\text{BiGDSSM-Block}^1_t(\cdot)$ are the spatial-temporal GDSSM Blocks of stream $1$, $x.\text{T}(a, b)$ denotes the transpose of $x$ and $x.\text{flatten}(a, b)$ the flattening operation of the $a$-th and $b$-th dimension of the tensor. The transpose, reshape and flattening operations are necessary to process the spatial and temporal dimension using the similar GDSSM blocks which expects a tensor of shape $B \times L \times D_m$.

The second stream aggregates information temporally and then spatially
\begin{align*}
 x_t &= \text{BiGDSSM-Block}_t^2(x.\text{T}(1,2).\text{flatten}(0,1))\\
 x_t &= x_t.\text{reshape}(B, J, F, D_m)\\
  x_{st} &= \text{BiGDSSM-Block}_s^2(x_t.\text{T}(1,2).\text{flatten}(0,1))\\
  x_{st} &= x_{st}.\text{reshape}(B, F, J, D_m)
\end{align*}
Finally, we combine the outputs of both the streams using learnable weights given by
\begin{align*}
    \begin{bmatrix}\alpha_{st}&\alpha_{ts}\end{bmatrix} &= \text{softmax}(\begin{bmatrix}x_{st} & x_{ts}\end{bmatrix}\mathcal{W})\quad \in \mathbb{R}^{B\times F\times J\times 2}\\
    x_{out} &= \alpha_{st} \odot x_{st} + \alpha_{ts} \odot x_{ts}\quad \in \mathbb{R}^{B\times F\times J\times D_m}
\end{align*}
where $\mathcal{W} \in \mathbb{R}^{2D_m\times 2}$ is a learnable mapping to the weights which are normalized by using $\text{softmax}(\cdot)$.

\paragraph{Causal model.} To design a causal variant of the spatiotemporal layer, we replace the temporal blocks in both the streams with an unidirectional GDSSM block as proposed in \cref{subsec:unigdssm}. In particular, we replace $\text{BiGDSSM-Block}_t^1(\cdot)$ $\text{BiGDSSM-Block}_t^2(\cdot)$ with $\text{UniGDSSM-Block}_t^1(\cdot)$ and $\text{UniGDSSM-Block}_t^2(\cdot)$.

\subsection{Pretraining HumMUSS}
\label{sec:pretrain}
To demonstrate HumMUSS ability to learn generic motion features, we pretrain our model on a pretext task, and then finetune it on downstream tasks that require human motion understanding. To ensure a fair comparison, we adopt the same pretraining strategy and datasets as MotionBERT~\cite{zhu2023motionbert}, which has achieved state-of-the-art results in various motion understanding tasks. We also pretrain causal variants of HumMUSS and MotionBERT~\cite{zhu2023motionbert}. Causal variant of MotionBERT is implemented by employing a causal mask to establish a robust baseline in the causal setup. In all our experiments, we use 16M\footnote{\cite{zhu2023motionbert} shows that a smaller 16M parameter version achieves comparable performance to the original 44M parameter model while being computationally cheap.} parameters for both models unless specified otherwise (see supplementary material for implementation details).

First, we aim to learn a robust motion representation using a universal pretext task. We employ a ``cloze'' task, akin to recovering depth information from 2D visual observations, inspired by 3D human pose estimation. We use large-scale 3D motion capture data such as the AMASS dataset~\cite{AMASS:ICCV:2019} to create a 2D-to-3D lifting task, where we generate corrupted 2D skeleton sequences from 2D projections of 3D motion. These sequences mimic real-world issues like occlusions and errors. Then, HumMUSS outputs motion representation which can be used to reconstruct 3D motion using a final linear layer. With the reconstructed and ground-truth (GT) 3D motion represented as $\hat{\mathbf{x}}$ and $\mathbf{x}$ respectively, the total loss in 3D space is given by
\begin{equation}
 \mathcal{L}_{3D} = \sum_{t=1}^{F} \sum_{j=1}^{J} \lVert \hat{\mathbf{x}}_{t,j} - \mathbf{x}_{t,j} \rVert^2 + \lambda \lVert \hat{\mathbf{v}}_{t,j} - \mathbf{v}_{t,j} \rVert^2   
\end{equation}
where $\hat{\mathbf{v}}_{t,j} = \hat{\mathbf{x}}_{t,j} - \hat{\mathbf{x}}_{t-1,j}$ and $\mathbf{v}_{t,j} = \mathbf{x}_{t,j} - \mathbf{x}_{t-1,j}$ are 3D velocities, and $\lambda$ is the weight of the velocity loss. For this pretraining task, we rely on two datasets with 3D GT motion trajectories: MPI-INF-3DHP~\cite{mehta2017monocular} and AMASS~\cite{AMASS:ICCV:2019}.  

Second, to utilize heterogeneous human motion data in various formats, we extract 2D skeletons from different motion data sources using in-the-wild RGB video datasets such as PoseTrack~\cite{8578640} and InstaVariety~\cite{kanazawa2019learning}. Given the dataset RGB videos, we either use the provided manually labelled 2D  GT skeletons, or follow~\cite{zhu2023motionbert} in computing them using the publicly available 2D pose estimation network from~\cite{newell2016stacked}. 
Since GT 3D motion is not available for this data, we use a weighted 2D re-projection loss. The final pre-training loss in the 2D image space is computed as
\begin{equation}
 \mathcal{L}_{2D} = \sum_{t=1}^{F} \sum_{j=1}^{J} \delta_{i,j}\lVert \hat{x}_{t,j} - x_{t,j} \rVert^2    
\end{equation}
with $\hat{x}$ the 2D orthographic projection of the predicted 3D motion $\hat{\mathbf{x}}$ and $\delta$ the 2D joint detection confidence.

To degrade the 2D skeletons, we apply random zero masking to $15\%$ of the joints and sample noises from a combination of Gaussian and uniform distributions~\cite{chang2019poselifter}. We use curriculum learning to pretrain HumMUSS for 90 epochs where $\mathcal{L}_{3D}$ is minimized for the first 30 epochs and $\mathcal{L}_{2D}$ is minimized in the remaining 60 epochs.

\begin{table}[!h]
\begin{center}
\scalebox{0.75}{
\begin{tabular}{lcc|ccc}
\hline
Method & $F$ & Params & MPJPE$\downarrow$ & PCK$\uparrow$ & AUC$\uparrow$ \\ \hline \hline
\multicolumn{6}{c}{Causal} \\
\hline
Mehta et al.~\cite{mehta2017monocular} & 1 & - & 117.6 & 75.7 & 39.3 \\ 
$^\dagger$MotionBERT (s)~\cite{zhu2023motionbert} & 243 & 16M & 25.4 & 97.9 & 85.2 \\
$^\dagger$MotionBERT (f)~\cite{zhu2023motionbert} & 243 & 16M & 21.1 & \textbf{99.0} & \textbf{86.8}\\
\hline
\rowcolor{lightgray}
HumMUSS (s) & 243 & 16M & 24.6 & 98.2 & 85.6 \\ 
\rowcolor{lightgray}
HumMUSS (f) & 243 & 16M & \textbf{21.0} & 98.7 & 86.1 \\ 
\hline \hline
\multicolumn{6}{c}{Bidirectional} \\
\hline
UGCN ~\cite{wang2020motion} & 96 & - & 68.1 & 86.9 & 62.1 \\ 
PoseFormer~\cite{zheng20213d} & 81 &  - & 77.1 & 88.6 & 56.4 \\ 
MHFormer~\cite{li2022mhformer} & 9 & 30.9M & 58.0 & 93.8 & 63.3 \\ 
MixSTE~\cite{zhang2022mixste} & 27 & 33.6M & 54.9 & 94.4 & 66.5\\ 
Einfalt \emph{et al.} \cite{einfalt2023uplift} & 81 & 10.4M &  46.9 & 95.4 & 97.6\\
P-STMO \cite{shan2022p} & 81 & 6.2M & 32.2 & 97.9 & 75.8\\
HDFormer \cite{chen2023hdformer} & 96 & 3.7M & 37.2 & 98.7 & 72.9\\
HSTFormer \cite{qian2023hstformer} & 81 & 22.7M & 41.4& 97.3 &71.5\\
STCFormer\cite{tang20233d} & 81 & 4.7M& 23.1 & 98.7 & 83.9\\
PoseFormerV2\cite{zhao2023poseformerv2} & 81 & 14.3M & 27.8& 97.9 & 78.8\\
GLA-GCN \cite{yu2023gla} & 81 & 1.3M & 27.7& 79.1 & 98.5 \\
MotionAGFormer \cite{mehraban2024motionagformer} & 81 & 19M & 16.2 & 98.2 & 85.3 \\
$^\dagger$ MotionBERT (s)~\cite{zhu2023motionbert} & 243 & 16M & 18.2 & 99.1 & 88.0 \\
$^\dagger$ MotionBERT (f)~\cite{zhu2023motionbert} & 243 & 16M & \textbf{16.0} & \textbf{99.3} & \textbf{89.9} \\
\hline
\rowcolor{lightgray}
HumMUSS (s) & 243 & 16M & 18.7 & 99.0 & 87.1 \\ 
\rowcolor{lightgray}
HumMUSS (f) & 243 & 16M & 16.3 & 99.2 & 89.2  \\ 
\hline
\end{tabular}
}
\end{center}
\caption{\textbf{Human 3D Pose Estimation}  Comparison on the MPI-INF-3DHP dataset. MPJPE (in $mm$) from detected 2D poses are reported. $F$ and Params denote the context/clip length and the number of parameters used by the method respectively. $^\dagger$ indicates results obtained from finetuning official implementation of MotionBERT~\cite{zhu2023motionbert}. (s) indicates models trained from sctach, (f) models finetuned after pre-training.}
\label{tab:3dhp_detailed}
\vspace{-1em}
\end{table}

\begin{table}[h]
\begin{center}
\setlength{\tabcolsep}{9pt}
\resizebox{\columnwidth}{!}{
\small
\begin{tabular}{lcc|ccc}
\hline 
Method & Input & $F$ & MPVE$\downarrow$ & MPJPE$\downarrow$ & PA-MPJPE$\downarrow$ \\
\hline \hline
\multicolumn{6}{c}{Causal} \\
\hline 
HMR \cite{kanazawa2018end}  & image & 1 & - & 130.0 & 81.3  \\
$^\dagger$ SPIN \cite{kolotouros2019learning}  & image & 1 & 129.1 & 100.9 & 59.1  \\
Pose2Mesh \cite{choi2020pose2mesh}  & 2D pose & 1 & 109.3 & 91.4 & 60.1 \\
I2L-MeshNet \cite{moon2020i2l}  & image & 1 & 110.1 & 93.2 & 58.6 \\
$^\dagger$ HybrIK \cite{li2021hybrik}  & image & 1 & 82.4 & 71.3 & 41.9  \\
METRO \cite{lin2021end} & image & 1 & 88.2 & 77.1 & 47.9 \\
Mesh Graphormer\cite{li2022mhformer}  & image & 1 & 87.7 & 74.7 & 45.6 \\
PARE \cite{kocabas2021pare}  & image & 1 & 88.6 & 74.5 & 46.5 \\
ROMP \cite{sun2021monocular}  & image & 1 & 108.3 & 91.3 & 54.9 \\
PyMAF \cite{zhang2021pymaf}  & image & 1 & 110.1 & 92.8 & 58.9 \\
ProHMR \cite{kolotouros2021probabilistic} & image & 1 & - & - & 59.8\\
OCHMR \cite{khirodkar2022occluded}  & image & 1 & 107.1 & 89.7 & 58.3 \\
3DCrowdNet \cite{choi2022learning}& image & 1 & 98.3 & 81.7 & 51.5 \\
CLIFF \cite{li2022cliff}  & image & 1 & 81.2 & 69.0 & 43.0 \\
FastMETRO \cite{cho2022cross}  & image & 1 & 84.1 & 73.5 & 44.6 \\
VisDB \cite{yao2022learning} & image & 1 & 85.5 & 73.5 & 44.9 \\
MotionBERT (f) \cite{zhu2023motionbert} & 2D motion & 16 & 93.5 & 82.3 & 50.9  \\
 MotionBERT (f)  + \cite{li2021hybrik}& video & 16 & 80.9 & 70.1 & \textbf{41.3}  \\
\hline
\rowcolor{lightgray}
HumMUSS (f) & 2D motion & 16  & 93.4 & 82.0 & 50.2   \\
\rowcolor{lightgray}
HumMUSS (f) + \cite{li2021hybrik} & video & 16 & \textbf{80.5} & \textbf{69.8} & \textbf{41.3} \\
\hline
\hline 
\multicolumn{6}{c}{Bidirectional} \\
\hline
TemporalContext\cite{arnab2019exploiting} & video & 32  & - & - & 72.2 \\
HMMR \cite{kanazawa2019learning}  & video & 20  & 139.3 & 116.5 & 72.6 \\
DSD-SATN\cite{sun2019human} & video & 9  & - & - & 69.5 \\
VIBE\cite{kocabas2020vibe}  & video & 16  & 99.1 & 82.9 & 51.9 \\
TCMR \cite{choi2021beyond}  & video & 16  & 102.9 & 86.5 & 52.7 \\
$^\dagger$ MAED \cite{wan2021encoder}  & video & 16  & 93.3 & 79.0 & 45.7 \\
MPS-Net \cite{wei2022capturing} & video & 16  & 99.7 & 84.3 & 52.1 \\
$^\dagger$ PoseBERT \cite{baradel2022posebert}  (+\cite{kolotouros2019learning}) & video & 16  & - & - & 57.3 \\
$^\dagger$ SmoothNet \cite{zeng2022smoothnet}  (+\cite{kolotouros2019learning}) & video & 32  & - & 86.7 & 52.7 \\
$^\dagger$ MotionBERT (f) \cite{zhu2023motionbert} & 2D motion & 16 & 88.1 & 76.9 & 47.2 \\
$^\dagger$ MotionBERT (f) + \cite{li2021hybrik} & video & 16 & \textbf{79.4} & \textbf{68.8} & \textbf{40.6} \\

\hline
\rowcolor{lightgray}
HumMUSS (f) & 2D motion & 16  & 88.9 & 77.4 & 47.5   \\
\rowcolor{lightgray}
HumMUSS (f) + \cite{li2021hybrik} & video & 16  & 80.0 & 69.1 & 40.7   \\
\hline 
\end{tabular}
}
\end{center}
\caption{\textbf{Human mesh recovery} Quantitative comparison on 3DPW dataset. Input and $F$ correspond to the input type and context length used by the method. $^\dagger$ denotes that the results are taken from ~\cite{zhu2023motionbert}. (s) indicates models trained from sctach, (f) models finetuned after pre-training.}
\vspace{-1em}
\label{tab:mesh_3dpw_only}
\end{table}

\section{Experiments}
\label{sec:experiments}
In this section, we evaluate pre-trained HumMUSS using different downstream tasks. Following ~\cite{zhu2023motionbert}, we choose 3D pose estimation, human mesh recovery and action recognition as our downstream tasks. After adding necessary heads on top of the HumMUSS backbone to produce outputs specific to each task, we finetune the entire model and showcase its competitive performance against current state-of-the-art models.
Finally, we demonstrate the benefits of using HumMUSS over Transformer-based methods in Sec~\ref{sec:hummus_vs_trasnf} and provide additional experiments in the supplementary material.

\subsection{3D Pose Estimation}

Since HumMUSS learns to predict 3D poses during pretraining, we directly finetune it using the final linear layer from the pretraining phase on the MPI-INF-3DHP dataset~\cite{mehta2017monocular}. Following previous work, we use the ground truth 2D skeletons from the videos in the dataset. In Table \ref{tab:3dhp_detailed}, we report the mean per joint position error (MPJPE) in millimeters (mm), Percentage of Correct Keypoint (PCK) within 150 mm range, and Area Under the Curve (AUC) as evaluation metric on the MPI-INF-3DHP dataset~\cite{mehta2017monocular}. We categorize the methods based on their causal or non-causal nature. Causal methods are more applicable to real-time scenarios where future frame information is unavailable. For a strong baseline we also finetune and train both causal and non-causal variant of MotionBERT on MPI-INF-3DHP.
We observe that HumMUSS consistently outperforms existing methods in the causal setup and competes favorably with state-of-the-art results in the bidirectional setup.

\subsection{Mesh Recovery}
We perform experiments on
3DPW~\cite{von2018recovering} datasets. Following prior work~\cite{zhu2023motionbert, zhang2022mixste, zheng20213d}, we augment the training set with the COCO~\cite{lin2014microsoft} dataset. In Table~\ref{tab:mesh_3dpw_only}, we report the performance of our finetuned model using MPJPE (mm), PA-MPJPE (mm) and MPVE (mm) metrics. Additionally, results of finetuning both the causal HumMUSS and MotionBERT for the Mesh Recovery task are also provided. Our model is competitive with existing approaches. 
However, as highlighted by \cite{zhu2023motionbert}, recovering full-body mesh solely from sparse 2D keypoints is inherently challenging due to the lack of human shape information. Therefore, we also provide results with a hybrid approach introduced by~\cite{zhu2023motionbert} that use an MLP to combine pretrained motion representations, and an initial prediction provided by RGB-based method HybrIK~\cite{li2021hybrik} to refine joint rotations. We notice that the hybrid approach significantly improves performance in both the causal and bidirectional setup making HumMUSS competitive with existing approaches. 


\begin{table}[h]
\begin{center}
\vspace{1em}
    \resizebox{0.75\linewidth}{!}{
        \begin{tabular}{lcc}
            \hline
            Method & X-Sub$\uparrow$ & X-View$\uparrow$ \\
            \hline
            ST-GCN~\cite{yan2018spatial} & 81.5 & 88.3 \\
            2s-AGCN~\cite{shi2019two}  & 88.5 & 95.1 \\
            MS-G3D~\cite{liu2020disentangling} & 91.5 & 96.2 \\
            Shift-GCN~\cite{cheng2020skeleton}  & 90.7 & 96.5 \\
            CrosSCLR~\cite{li20213d}  & 86.2 & 92.5 \\
            MCC (finetune)~\cite{su2021self} & 89.7 & 96.3 \\
            SCC (finetune)~\cite{yang2021skeleton}  & 88.0 & 94.9 \\
            UNIK (finetune)~\cite{yang2021unik} & 86.8 & 94.4 \\
            CTR-GCN~\cite{chen2021channel} & 92.4 & 96.8 \\
            PoseConv3D~\cite{duan2022revisiting} & 93.1 & 95.7 \\
            $^\dagger$ MotionBERT (finetune)~\cite{zhu2023motionbert} & 93.0 & 97.2\\
            UPS~\cite{foo2023unified} & 92.6 & 97.0 \\
            SkeleTR [C]~\cite{duan2023skeletr} & \textbf{94.8} & \textbf{97.7} \\
            \hline
            \rowcolor{lightgray}
            HumMUSS (finetune) & 92.0 & 97.4 \\
            \hline
        \end{tabular}
    }
    \caption{\textbf{Action Recognition} Quantitative comparison on the NTU-RGB+D dataset. Left and right column report the top-1 accuracy for the cross-subject and cross-view split respectively. $^\dagger$ indicates results taken from ~\cite{zhu2023motionbert} using the 44M parameter version of MotionBERT.}
    \label{tab:action}
    \vspace{-2em}
\end{center}
\end{table}

\subsection{Skeleton-based Action Recognition}
For this task, following \cite{zhu2023motionbert}, we perform global average pooling across motion representations of different persons and timesteps. The outcome is subsequently input into an MLP with one hidden layer. The network is trained using cross-entropy classification loss. In Table~\ref{tab:action}, we present a comparison between HumMUSS and recent SOTA methods for action recognition on the NTU-RGBD dataset~\cite{liu2019ntu}. Following~\cite{zhu2023motionbert}, we finetune pretrained HumMUSS on the training set of NTU-RGBD. We observe that HumMUSS performs favorably compared to recent methods and is only outperformed by SkeleTR~\cite{duan2023skeletr}, the most recent approach to action recognition. 
It is worth noting that while SkeleTR~\cite{duan2023skeletr} employs an extremely task specific architecture, HumMUSS is designed to serve as a generic motion understanding backbone that can be applied to several other motion-related tasks. Enhancing performance through modifications 
in the fine-tuning architecture is left as future work.


\subsection{HumMUSS vs Transformer-based methods}
\label{sec:hummus_vs_trasnf}
HumMUSS offers several advantages over existing SOTA models that depend on transformer-based architectures. In this section, we delve into the benefits of HumMUSS over one such method, MotionBERT~\cite{zhu2023motionbert}.\footnote{We use same number of parameters (16M) for both the models.}

\begin{figure}[h]
\centering
    \includegraphics[width=\columnwidth]{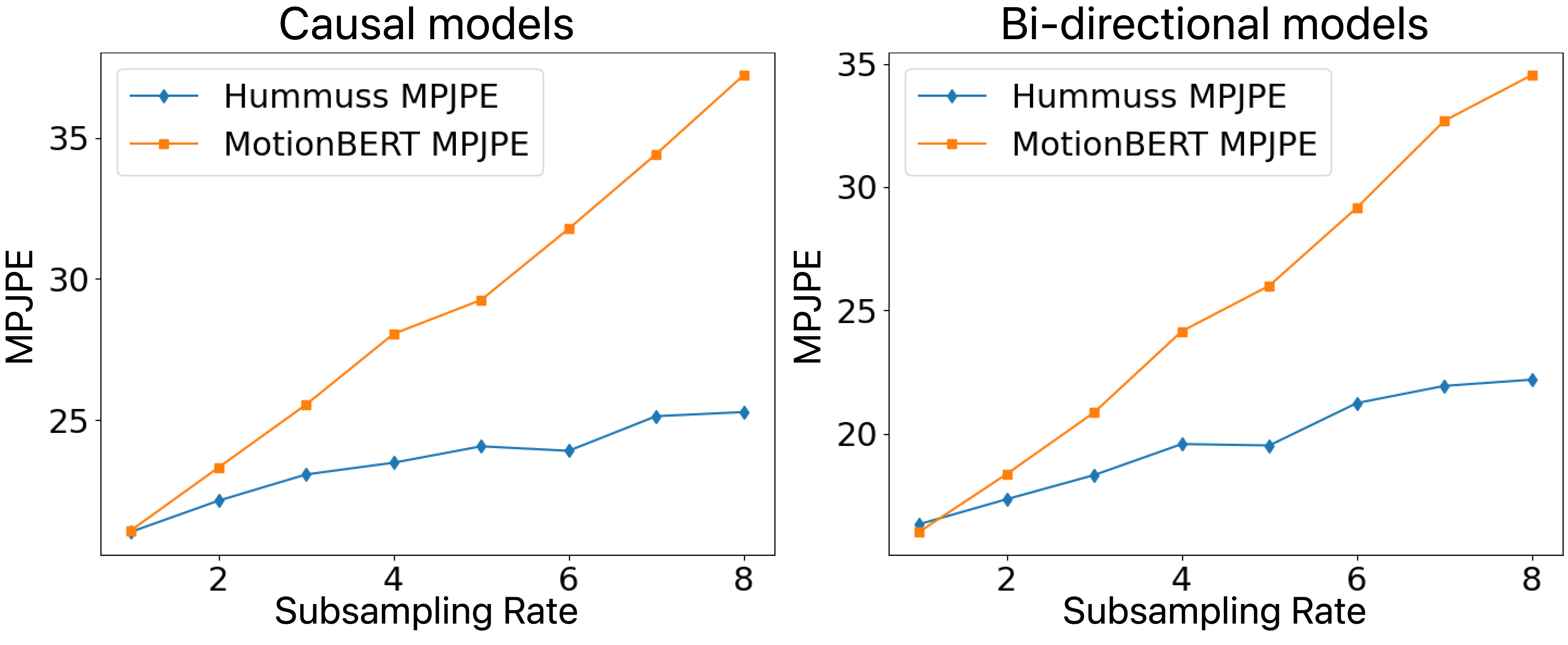}
\caption{Comparison between HumMUSS and MotionBERT~\cite{zhu2023motionbert} 3D pose estimation performance (MPJPE in $mm$) on MPI-INF-3DHP at different sub-sampling rates. 
\textit{Left}: causal models; \textit{Right}: bi-directional models.}
\label{fig:sampling_plot}
\end{figure}

\begin{figure}[h]
\centering
    \includegraphics[width=\columnwidth]{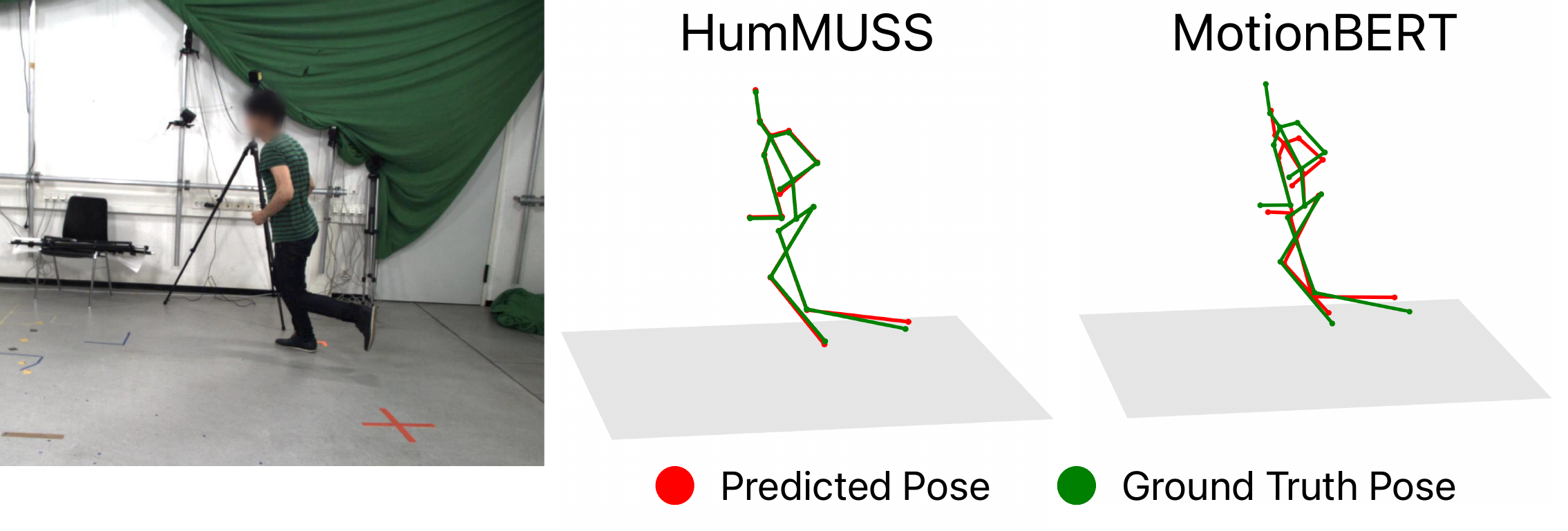}
\caption{Example of reconstructed 3D poses between HumMUSS and MotionBERT~\cite{zhu2023motionbert} when input signal is sub-sampled at rate 8 (using one frame every 8). Ground-truth represented in green, predictions in red. Best viewed in color.\vspace{-2mm}}
\label{fig:comparison_sampling}
\end{figure}

\textbf{Robust to new frame rate.}
In contrast to transformer-based architectures, HumMUSS, being a continuous-time model, demonstrates the ability to generalize to unseen frame rates with minimal drops in accuracy. This adaptability can be achieved by appropriately adjusting the discretization parameter $\Delta$ within the model using the new frame rate. Figures \ref{fig:sampling_plot} and \ref{fig:comparison_sampling} present both qualitative and quantitative comparison between HumMUSS and MotionBERT \cite{zhu2023motionbert} trained on the MPI-INF-3DHP dataset \cite{mehta2017monocular} and evaluated on sub-sampled motion videos from the test set. The findings reveal that transformer-based methods like MotionBERT's performance declines significantly as the sub-sampling rate increases compared to HumMUSS. 
The capability to perform reliably under various sampling rates is valuable in real-world scenarios, where input frame rates might fluctuate due to factors such as thermal throttling of the capturing devices. We provide a more detailed discussion in the supplementary material.

\begin{figure}[h]
\centering
    \includegraphics[width=\columnwidth]{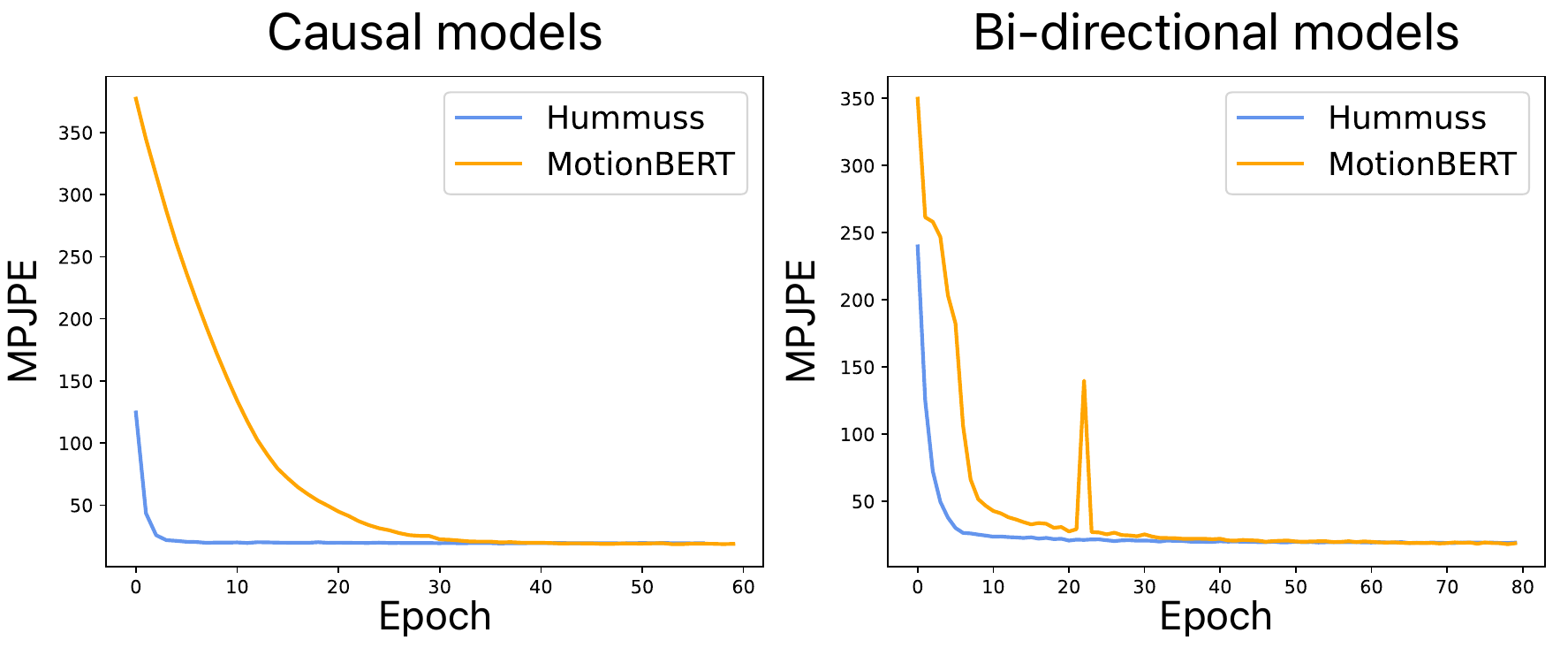}
\caption{Training dynamics of HumMUSS and MotionBERT~\cite{zhu2023motionbert} for 3D pose estimation on MPI-INF-3DHP. We plot the validation performance (MPJPE in $mm$) over epochs.  
\textit{Left}: causal models; \textit{Right}: bi-directional models.}
\label{fig:training_dynamics}
\end{figure}

\textbf{Training Speed and convergence} 
The training speed of HumMUSS scales favorably due to its \( O(F \log(F)) \) complexity, in contrast to the \( O(F^2) \) complexity of attention-based architectures, where $F$ represents the number of frames. As a result, HumMUSS is significantly faster than MotionBERT for longer sequence lengths. Training with longer context length can be crucial for certain human motion understanding tasks including action recognition, gesture analysis and Gait Analysis. Furthermore, HumMUSS demonstrates significantly faster convergence compared to MotionBERT during the training phase both for the bidirectional and causal model, as illustrated in \cref{fig:training_dynamics}. Exploring the underlying reasons for these training dynamics presents an interesting avenue for future research.



\begin{figure}[h]
\centering
    \includegraphics[width=0.9\columnwidth]{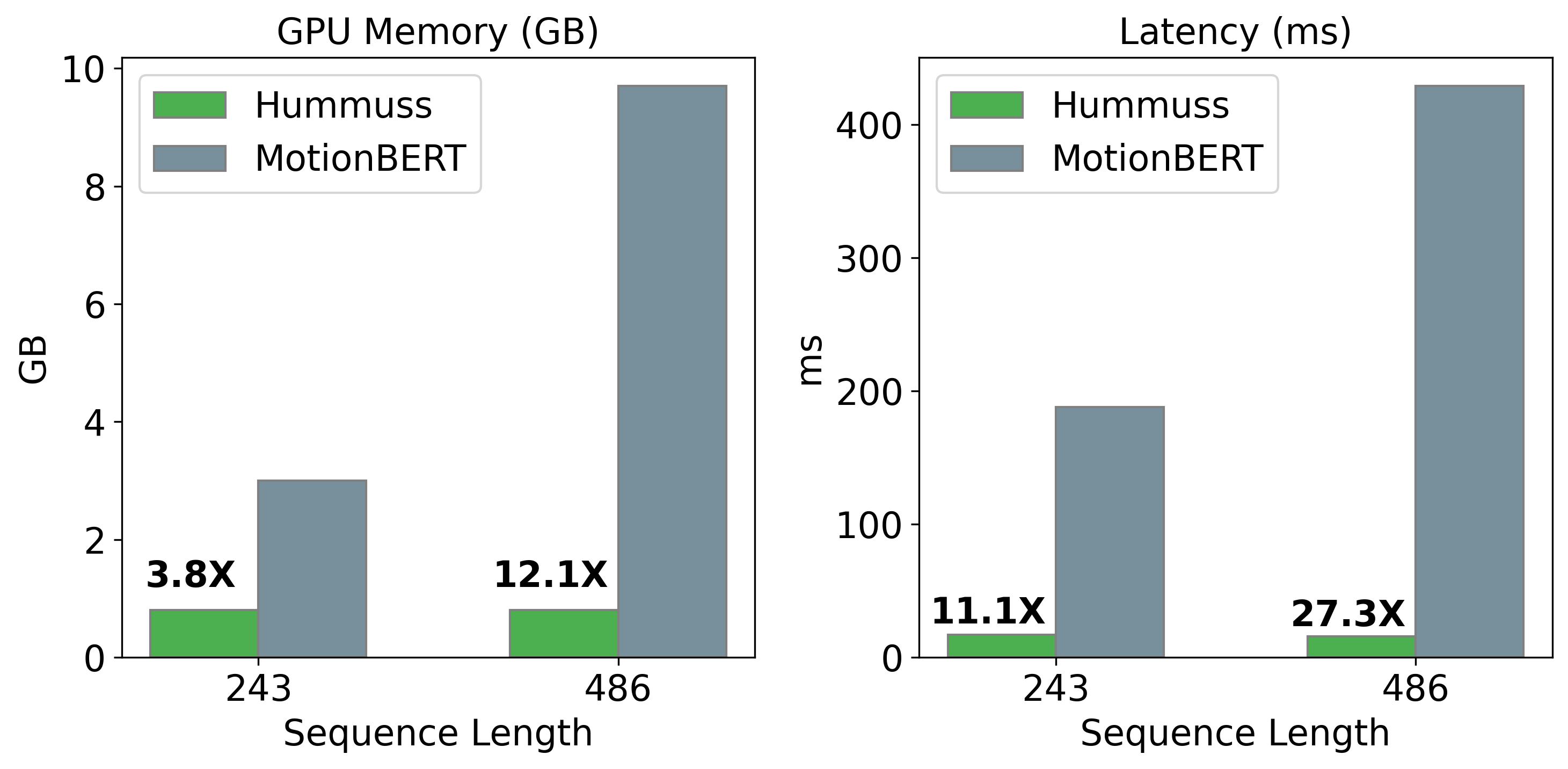}
\caption{GPU memory usage and Latency comparison between MotionBERT and HumMUSS for sequential inference using a causal model. Both the models are tested on a 80GB A100 GPU with a batch-size of 32. GPU memory utilized is reported in GB and average inference latency is reported in ms for every frame.}
\label{fig:comparison_inference}
\end{figure}

\textbf{Efficient Sequential Inference.}
One of the remarkable advantages of HumMUSS lies in its efficiency for real-time sequential inference, making it a drop in replacement for transformer-based models in various high-accuracy real-time applications. HumMUSS operates as a stateful recurrent model during sequential inference, requiring only the current frame and the state that summarizes the past frames. This substantially boosts the inference speed and efficiency of HumMUSS relative to MotionBERT~\cite{zhu2023motionbert}\footnote{We used the official implementation of MotionBERT. Adding causal transformer inference speedup tricks like KV-caching will improve MotionBERT's speed and memory. }. As depicted in \cref{fig:comparison_inference}, we observe HumMUSS is $3.8\times$ memory efficient and an $11.1\times$ faster during sequential inference on $243$ frames. Moreover, the constant memory usage and inference speed of HumMUSS result in increasingly significant improvements in GPU memory and latency for longer frame sequences.

\section{Conclusion}
\label{sec:experiments}

In this work, we present HumMUSS, a novel attention-free architecture designed specifically for human motion understanding. HumMUSS leverages diagonal state space models, effectively overcoming some of the major limitations inherent in current state-of-the-art transformer-based models, notably their slow sequential inference and limited robustness when faced with various frame rates.
Our extensive experiments highlight HumMUSS' versatility in a range of motion understanding tasks, proving its effectiveness in 3D pose estimation, mesh estimation, and action recognition. We are confident that HumMUSS can serve as a powerful motion understanding backbone, especially in applications that require real-time sequential inference. This approach marks a significant step forward, potentially leapfrogging existing methods and narrowing the divide between highly accurate transformer-based techniques and their practical application in the real world.
{
    \small
    \bibliographystyle{ieeenat_fullname}
    \bibliography{main}
}
\maketitlesupplementary
\section{Implementation Details}
\label{appdx:implement_details}







Our pretraining approach builds on top of open-sourced implementation of MotionBERT~\cite{zhu2023motionbert} \footnote{https://github.com/Walter0807/MotionBERT} . We implement our model using PyTorch and adapt the data-preprocessing pipeline from MotionBERT for both pretraining and finetuning phases. Detailed information about this pipeline can be found in \cite{zhu2023motionbert}. The experiments were conducted on a linux machine equipped with 4 NVIDIA A100 80GB GPUs. However, for finetuning experiments, a single A100 80GB GPU proved to be sufficient. 

HumMUSS is designed with $5$ spatiotemporal blocks, that is \(N = 5\). Given that we work with 2D joints and their confidences, the input dimensionality is set to \(D_{in} = 3\). Our implementation utilizes a \(16M\) parameter model for HumMUSS, with a model dimension (\(D_m\)) of $256$.

In our approach, the hyperparameters for GDSSM Blocks are adjusted to make both the causal and unidirectional model consist of $16M$ parameters. Specifically, we use \(k=1\) for the spatial GDSSM block and \(k=2\) for the temporal GDSSM block. For the causal HumMUSS model, we set \(m=3\) for both the spatial and temporal blocks, while for the bidirectional model, \(m=2.5\) is used. Furthermore, the dimension of the state space (\(N\)) is set to $128$ for both variants of HumMUSS, and the representation dimension (\(D_{rep}\)) is $512$.

Apart from the learning rate, the other training hyperparameters for both the pretraining and finetuning tasks are kept consistent with those used in the MotionBERT model \cite{zhu2023motionbert}, ensuring a fair comparison with the transformer-based baseline. We use a learning rate of 0.0003 for the pretraining. We use similar learning rates as \cite{zhu2023motionbert} for finetuning tasks. 

\begin{figure*}[tb]
    \centering
    \includegraphics[width=\linewidth]{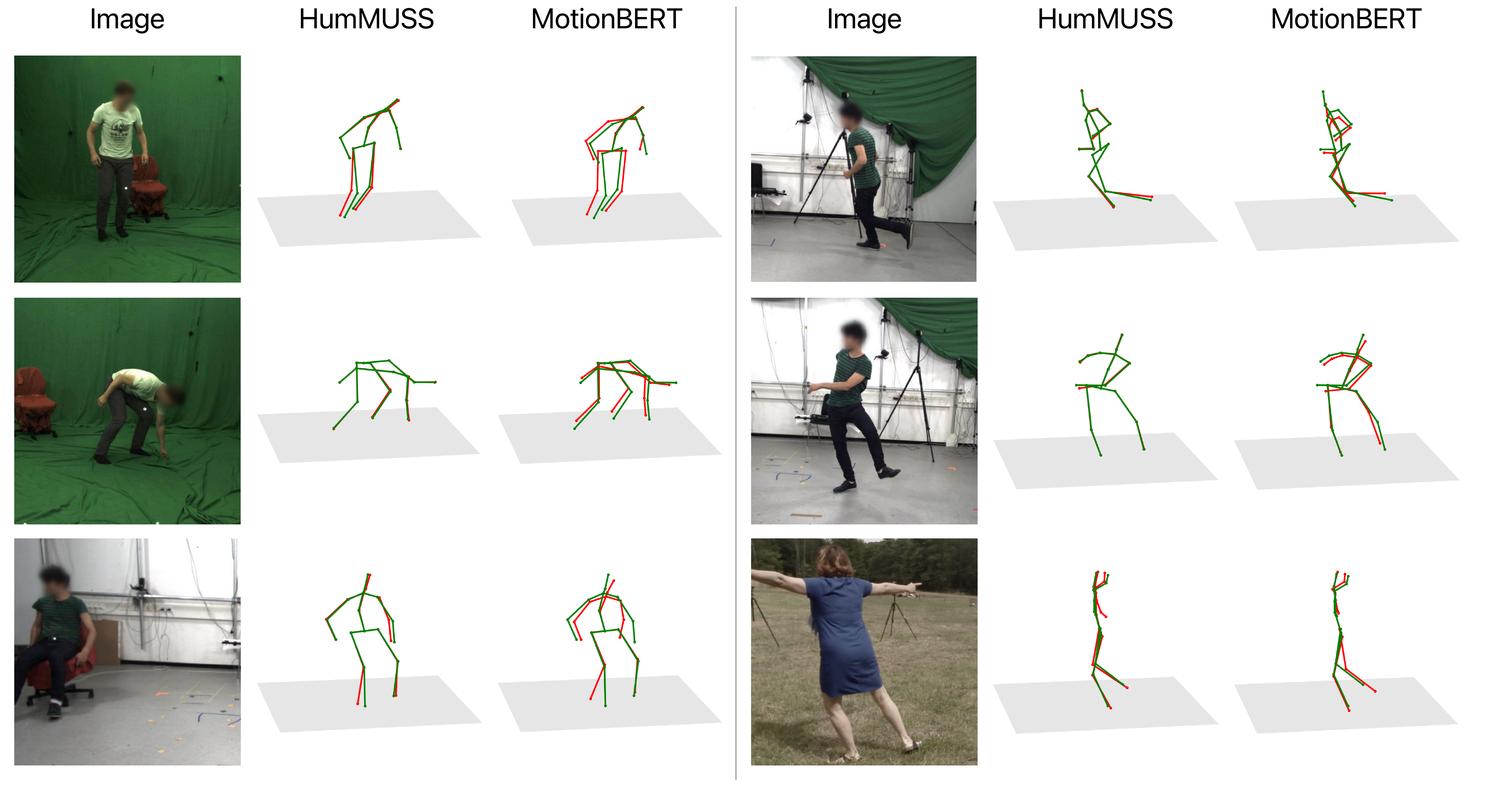}
    \caption{Example frames of 3d reconstruction for both HumMUSS and MotionBERT~\cite{zhu2023motionbert} when the input signal is sub-sampled by a factor of 8. Green and red poses represent \textit{ground truth} and \textit{prediction} respectively. Best viewed in color.}
    \label{fig:qualitative_res}
\end{figure*}

\section{Additional experiments}
\label{appdx:additional_experiments}

In this section, we delve deeper with additional experiments to explore the impact of architectural choices on HumMUSS. These investigations are specifically conducted within the domain of 3D Pose Estimation, utilizing the MPI-INF-3DHP dataset \cite{mehta2017monocular}. This task choice aligns well with our primary goal of pretraining objective to learn motion representations.

\subsection{Generalization to longer context window}

We show the ability of HumMUSS's architecture to generalize to longer sequence lengths in this experiment. The model, initially trained on a sequence length of F=27 frames, was tested on longer sequences of F=81, and F=243. As shown in Table 2, the HumMUSS model exhibits a consistency in performance when tested on longer sequences. This means HumMUSS can generalize to longer sequences than the one it is trained while being able to encode the longer context. This proves especially beneficial for deployment in tasks necessitating a longer contextual understanding than what was used during the model's training. 

\begin{table}[!h]
\begin{center}
\resizebox{\linewidth}{!}{
\begin{tabular}{c|ccc}
\hline
 Training sequence length & & Test sequence length & \\
 F=27 & F=27 & F=81 & F=243 \\
\hline
HumMUSS    & $25.1$   & $25.8$  & $26.2$      \\
\hline
\end{tabular}}
\caption{3D Pose Estimation performance (MPJPE in $mm$) of a HumMUSS model trained on 27 frame and tested on longer sequences.}
\label{tab:perf_seq_generalization}
\end{center}
\end{table}

\subsection{Ablation study}

An ablation study was conducted to evaluate the impact of different aggregation methods in the GDSSM (Generalized Dynamic Spatiotemporal Sequence Modeling) blocks of the HumMUSS model. In particular, we consider aggregation of the different information processing pathways in GDSSM blocks. In Section~4.1 and~4.2, we use Multiplicative gating to combine $x_f$, $x_b$ and $x_id$. In this ablation, we also test other aggregation methods including average pooling and pooling with learnable weights. Notably, the pooling with learnable weights technique is the same as the one presented in Section~4.3 the results, as illustrated in Table 3, show that Multiplicative Gating outperforms other methods in both causal and bidirectional model types, with the lowest MPJPE of 40.2 mm in the bidirectional setting and 45.1mm in the causal setting. This indicates the effectiveness of Multiplicative Gating in aggregating information within GDSSM blocks.

\begin{table}[!h]
\begin{center}
\resizebox{0.9\linewidth}{!}{
\begin{tabular}{c|cc}
\hline
 Type of aggregation & \multicolumn{2}{c}{Model type}\\
 in GDSSM Blocks & Causal & Bidirection\\
\hline
Average Pool    & $25.8$   & $20.5$  \\
Pooling with learnable weights & $25.3$& $20.1$\\
Multiplicative Gating    & $24.6$   & $18.7$  \\
\hline
\end{tabular}}
\caption{3D Pose Estimation performance (MPJPE in $mm$) of HumMUSS using different aggregation methods to combine information in the GDSSM blocks.}
\label{tab:runtime}
\end{center}
\end{table}

Further, the performance of HumMUSS using different aggregation methods to combine two streams of Spatiotemporal GDSSMs was evaluated. The methods included Multiplicative Gating, Average Pooling, and Pooling with Learnable Weights. The findings, presented in Table 4, indicate that Pooling with Learnable Weights provides the best performance in both causal and bidirectional settings. This highlights the effectiveness of learnable weights in combining multiple streams of spatiotemporal data. Surprisingly, multiplicative gating performs poorly for combining two streams which warrants further investigation in the future. 

\begin{table}[!h]
\begin{center}
\resizebox{0.9\linewidth}{!}{
\begin{tabular}{c|cc}
\hline
 Type of aggregation & \multicolumn{2}{c}{Model type}\\
 to combine two streams & Causal & Bidirection\\
\hline
Multiplicative Gating    & $26.6$   & $21.1$  \\
Average Pool    & $25.1$   & $19.5$  \\
Pooling with learnable weights & $24.6$& $18.7$\\
\hline
\end{tabular}}
\caption{3D Pose Estimation performance (MPJPE in $mm$) of HumMUSS using different aggregation methods to combine the two streams of Spatiotemporal GDSSMs.}
\label{tab:runtime}
\end{center}
\end{table}

\section{Initialization of DSSM parameters}
\label{appdx:init_dssm}
A simple way to initialize the diagonal parameters is to use the linear initialization which sets the real part $\Lambda_{re}$ to $-\frac{1}{2}\mathbbm{1}_N$ and the imaginary part $\Lambda_{im}$ to $(\pi j)_{j=1}^N$~\cite{gu2022parameterization}. Gu~\etal~\cite{gu2022parameterization} provides many other initialization techniques and empirically verify that they can approximate the initialization of the S4~\cite{gu2021efficiently} kernel which leads to stable training regime for longer sequences. In practice, we found all the techniques leads to similar performance and use the simple linear initialization scheme throughout our experiments. Finally, elements of $C$ are samples from a normal distribution and $\Delta$ is initialized randomly between $0.001$ and $0.1$. 

\section{Generalization to unseen frame rate}
\label{appdx:unseen_frame_rate}
Transformers, being discrete models, are inherently designed for sequences that have fixed uniform spacing in time and utilize positional encodings to encode temporal information. MotionBERT, in particular, exacerbates this issue by employing learnable positional encodings, which lack a clear sense of time. 
In contrast, HumMUSS being a continuous time model, does not require positional encodings and enables training and testing on data with different frame rates. Unlike Transformers, which struggle with down-sampled or up-sampled videos, HumMUSS seamlessly adapts. For instance, if trained on 30 FPS data, changing the discretization parameter $\Delta$ to $2\Delta$ or $\frac{1}{2}\Delta$ allows the model to perform effectively on 15 or 60 FPS videos respectively, without the need for expensive re-training.

In practice, sampling rate changes can directly be measured by e.g. reading camera timestamps to detect frame drops or frame-rate changes due to external factors. As a continuous time model, HumMUSS interprets sequences as signal values at different timestamps. Adjustment the discretization parameter allows HumMUSS to process it as changes in the number of samples, rather than a different input signal.
Furthermore, in sequential inference scenarios with varying sampling rate, we keep HumMUSS robust by dynamically scaling $\Delta$ based on the evolving sampling rate. This is possible because HumMUSS maintains a compressed memory of the past observed signal in its state that is independent of past sampling frequencies. Such adaptability positions HumMUSS as a potent model for real-time applications where input frame rates may vary due to factors like thermal throttling of capturing devices.

\end{document}